# Automatic Guided Vehicles System and Its Coordination Control for Containers Terminal Logistics Application

Roni Permana Saputra, Estiko Rijanto
Research Centre for Electrical Power and Mechatronics, Indonesian Institute of Sciences,
Komplek LIPI, Jl. Cisitu No. 21/154D, Bandung 40135, Indonesia
permana.saputra@yahoo.co.id; estikorijanto@lipi.go.id

*Abstract* - **Automatic Guided Vehicle (AGV) has been widely applied in automatic logistics system because it provides flexibility and efficiency. This paper addresses review and design of multi AGVs system (MAGS) and its coordination control to be implemented at containers terminal logistics systems. It focuses on design of electric AGV and supervisory control to coordinate multi AGVs. Firstly, related previous works by other researchers are reviewed including containers terminal system, application of AGV in logistics systems and multi mobile robots control systems. Secondly, application of AGV system in containers terminal is proposed including AGV design specification, AGV control method, multi AGV system supervisory control method, and containers terminal port model. Conclusions have been obtained as follows: an electric AGV may be developed using electric motors with total power between 700-900KW and speed of 228 rpm; a MAGVS is proposed based on free ranging navigation system using GPS, digital compass and rotary encoder sensors; path planning, coordination and traffic management control is conducted by the supervisory controller based on priority based coordinative (PBC) control algorithm; a model of MAGVS for containers terminal logistics system can be developed using the navigation system and PBC control algorithm.**

## I. INTRODUCTION

Development of global trade has been growing rapidly which is characterized by a necessity of export and import activities inter-island and inter-state. To support these activities, logistic systems are very crucial in trade and distribution of goods between islands and states. Logistics system is becoming a potential business area so that it plays a decisive role as parameter of a nation's competitiveness [1]. To enhance national economy competitiveness, a qualified national logistics system (SISLOGNAS) becomes inevitable [2].

Meanwhile, according to data provided by the Indonesian Chamber of Commerce (KADIN), logistics cost in Indonesia is still considered being high when compared to Japan which is only 5% from total business costs. In the ASEAN region, Indonesia is still under the Philippines whose logistics cost is 7%, Singapore 6%, and Malaysia 8% of total business costs [3]. An international survey result shows that Indonesia logistics rank is quite apprehensive. Based on the World Bank survey, for the Logistics Performance Index, Indonesia ranks 75th of 155 countries, lays under other ASEAN countries, namely: Singapore (2), Malaysia (29), Thailand (35), Philipines (44), and Vietnam (53) [4]. In table 1, it is shown the logistics performance scores and rankings of Indonesia in several aspects.

TABLE I
INDONESIAN LOGISTICS PERFORMANCE INDEX [4]

|  | | **Indonesia** |
|---|---|---|
| Overall LPI | Score | 2.76 |
|  | Rank | 75 |
| Custom | Score | 2.43 |
|  | Rank | 72 |
| Infrastructure | Score | 2.54 |
|  | Rank | 69 |
| International shipments | Score | 2.82 |
|  | Rank | 80 |
| Logistics competence | Score | 2.47 |
|  | Rank | 92 |
| Tracking & tracing | Score | 2.77 |
|  | Rank | 80 |
| Timeliness | Score | 3.46 |
|  | Rank | 69 |

Indonesia logistics cost can account for 24% of total Gross Domestic Product (GDP) while Malaysia is only 15%. Furthermore, either in USA or Japan, logistics cost accounts for only 10% of total GDP [5].

Logistic cost encompasses containers loading-unloading cost at ports and transportation cost, and it is affected by time interval. For example to dismantle a 20-foot container loaded in Tanjung Priok port, it needs cost of 95 U.S. dollars, while in Malaysia is only 88 dollars, and in Thailand is only 63 U.S. dollars [5]. It requires 5.5 days for goods to be imported into Indonesia. According to the chairman of the Association of Logistics and Forwarder Indonesia (ALFI) [2], one reason of inefficiency of the national logistics industry is because Indonesia does not have a qualified logistics system with appropriate international standards in general. The process of both of containers loading-unloading and documents distribution still use conventional systems.

One way to reduce logistics cost in Indonesia is by improving logistic system in harbors. Automation technology in containers loading-unloading process may be adopted [6]. Technology that is quite commonly used in a harbor is the use of AGV system to transport containers from the shipping area to the container yard and vice versa. By using AGV system, containers transport can be set centrally and be more regular basis that yields shorter time interval. As a result, operating cost reduction can be obtained [7].

An AGV normally means a mobile robot which is used for transporting objects. They were traditionally employed in manufacturing systems, but have recently extended their popularity to many other industrial applications. AGV system was first introduced in 1955 for transporting materials. The use of AGVs for transporting containers was firstly applied in 1993 at the containers terminal in Rotterdam [7] [8] [9].

One important aspect in an AGV system is its control system. Traditionally, AGVs used in transportation tasks are controlled by a central server via wireless communication [9]. Along with demanding application which needs more flexibility and openness of system, researchers began to develop various control methods. For instance Danny Weyns et. al. developed an innovative decentralized architecture software for controlling AGVs [9]. Lawrence Henesey et. al. [10] introduces a new generation of AGVs which were evaluated in operations of a container terminal. Mohd Safwan [11] in his thesis focused on development of an AGV control system involving how an AGV operates, its movement and loading-unloading mechanism. One of recent technology that can be implemented is a multi-agent collaborative technology system which involves a group of agents in which each member behaves autonomously to reach the common goal of the group [12]. This technology provides high flexibility and effectiveness.

The main purpose of this paper is to review AGV systems and to propose a hypothetical design of electric AGV and its control system as well as container terminal port model.

## II. METHODOLOGY AND RELATED WORKS

In order to achieve the above purpose, in this paper the following procedure has been carried out:
1) Literature survey concerning containers terminal system.
2) Literature survey concerning application of AGV in logistics systems.
3) Literature survey concerning multi mobile robots coordination control system.
4) Proposing hypothetical design of AGV control system and containers logistics system modeling.

In the following, related previous works done by other researchers are reviewed.

### A. Container Terminal System

Container terminal is a place for loading and unloading of containers before they are distributed to the assigned destinations [13]. In general, based on distribution flow of containers, container terminal activities are divided into three categories namely [6]:
(a) export activity, in which a container arrives and enters into the container terminal by using a trailer or a train, then after transit, it is carried out leaving the terminal by a container ship,
(b) import activity, in which a container enters into the harbor using a container ship, then after transit in the container terminal, it is carried out leaving the terminal by a truck or a train, and
(c) transit activity, where a container is intended to transit temporary. It enters into the harbor using a container vessel, then after having transited in the terminal, it is then transported back to the next ship.

The third activity is common in a container terminal. Figure 1 shows illustration of activities and processes in a transit container terminal.

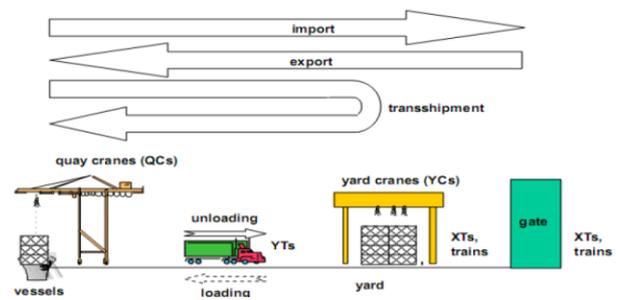

Fig. 1. Illustration of activities in a transit container terminal [6].

In general, process of loading and unloading of containers in a container terminal consists of three separate processes as follows [10].
1) Retrieval of containers from stacking lane
2) Transporting of containers from stack to quay
3) Transfer of containers from quay side to sea vessel

### B. Application of AGV in Logistic Systems

AGV system is widely used in logistics systems [14] [15]. Technological development related to AGV systems which include vehicle system, vehicle control, navigation, positioning, and communication with supervisory control is explained in paper [16]. Material handling system in a manufacturing unit and elements which are important in design of an AGV are addressed in paper [17]. An AGV navigation system may consist of a data collection, decision making, and hardware control systems. Its navigation method can be developed using fuzzy logic based neural network where angle and distance of an obstacle is measured to generate speed and steering angle [18].

In thesis [19] positioning control system of AGVs is presented. An AGV is usually a mobile robot that follows a specific routing of wire or magnetic tape, fast-flowing wire, or line, so that it can move freely without any guidance infrastructure. A system that can locate the robot in the form of absolute position or relative position is needed. The measurement technique is based on measuring relative position of the mileage accumulated by the robot relative to the starting position. The weakness of this method is that it allows error drift every time. Meanwhile, a digital compass or GPS is an example of absolute positioning technique in which a robot direction angle or position is determined from a single reading. Absolute positioning drawback is that it costs higher and it is system case where signal blockage may occur due to buildings. To overcome the shortcoming of each method, a method is demonstrated using a combination of absolute and relative sensors. Publication [20] demonstrates design and implementation results of RF-based wireless control to control a distributed AGVs. Paper [21] presents architecture of decentralized control of AGV to meet flexibility and openness where the software is structured as a multi agents AGV system.

Applications of AGVs in container terminals can be found in [22], [23] and [24]. Paper [22] describes design and control of AGVs system with a focus on application of container transport in a dock container terminal. Its case study focuses on control zone, path strategy, guide layout, traffic control and routing algorithms to minimize the AGV travel distance. Paper [23] explains AGV application for two containers yards. Whereas paper [24] presents a challenging coordination task for AGVs in a container terminal in order to identify the number of AGVs required to operate cranes with the principle of "cranes-no waiting" for loading and unloading containers to and from ships.

*C. Multi Robots Coordination Control System*

A variety of multi robots system related research has been reported by researchers. Various aspects are already widely reviewed in many articles. Most of them discuss about some aspects related to robot/vehicle control, trajectory planning, coordination algorithms, navigation systems, communication systems, etc [25]. Reference [26] presents observation algorithm of multiple moving targets in a distributed multi robots system. In this research cooperation in a multi robots team based on sensors is used for observation of multiple moving targets. Paper [27] concerns on developing on-line distributed control strategies that allow a robot team to attempt to minimize the total time. Paper [28] argues about coordinated motion control for multiple mobile robots where there exist obstacles along the way. The approach taken is intended to regulate the speed profile to be continuous in all the robots to avoid possible collisions at each intersection of two lines or more along the path specified. Meanwhile, paper [29] discusses design of multiple robot motion using priority based method. In this paper two methods of determining path are compared those are the coordinated multi robot method and the priority based method. Motion design based on priority method is a simple approach introduced by That Erdmann and Lozano [30].

## III. MAGVs SYSTEM FOR APPLICATION IN A CONTAINERS TERMINAL LOGISTICS SYSTEM

This paper proposes a multi AGV system (MAGVS) to be implemented at a container terminal logistics system to improve performance. The overall design specification of this MAGVS is as follows.
1) Capability to transport various types of standard containers which have standard weight and dimension.
2) Capability to know and to map the location of its working space without using of guidance in the form of electric wires, magnetic tape, lines and so forth, so that its movement can be more flexible.
3) Having ability to determine the shortest routing to pursue goals that are ordered by the supervisory control computer from the MAGVs initial position.
4) Being able to overcome static obstacles in their work environment and dynamic obstacles which are the movement of other AGVs so that there will be no collisions between AGVs and obstacles.
5) Being able to set the transportation priorities by giving priority path that is free from barriers to dynamic obstacle for the AGV which transports container.
6) Provides a good interaction between the supervisor computer and the MAGVs team.

To realize the above over all specifications, in the following detailed specifications are presented.

*A. The Proposed AGV Design Specification*

Figure 2 shows an example of AGV system used in container transport applications which is product of Gootwald Port Technology GmbH [31]. Some technical specifications of the AGV are shown in table 2. This AGV system uses diesel engine propulsion.

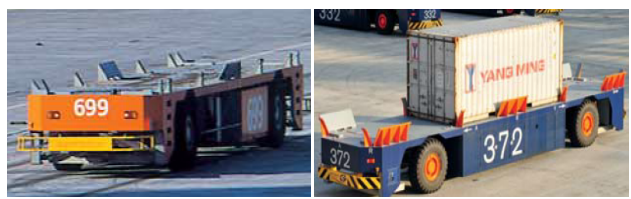

Fig. 2. An AGV system for containers transport applications [31]

Instead of using diesel engine propulsion, this paper proposes a MAGVS using electric motor drives with a view to provide an environmentally friendly MAGVS. In addition, using electric motor drives may allow faster maneuvers of MAGVS.

Some aspects must be considered in designing a MAGVS to meet the overall specification including: electric motor power, battery energy capacity, and steering control.

TABLE 2
AGV SPECIFICATION FOR CONTAINERS TERMINAL APPLICATION [31]

| Engine | |
|---|---|
| Diesel Engine Power | 1,200 HP |
| Fuel Tank Capacity | 1,400 L |
| **Container Types** | |
| | 1 x 20'; 1 x 40'; and 1 x 45' container |
| | 2 x 20' containers |
| | 1 x 30' container as an option |
| **Load Weights** | |
| Max Weight of single container | 40 ton |
| Max Weight of 2 x 20' container | 60 ton |
| **Dimensions** | |
| Length | approx. 14.8 m |
| Width | approx. 3.0 m |
| Loading area height | approx. 1.7 m |
| Dead weight | approx. 25 ton |
| Tyre size | 18.00 R 25 |
| **Speed & Acceleration** | |
| Max. speed forward/reverse | 6 m/s |
| Max. speed in curves | 3 m/s |
| Max. crab steering speed | 1 m/s |
| Max. acceleration | 2 m/sec² |

*Electric Motor Power Requirement*

Appropriate electric motor is required for an AGV so that it has sufficient power to transport standard containers. Figure 3 shows a model of AGV which is transporting a container.

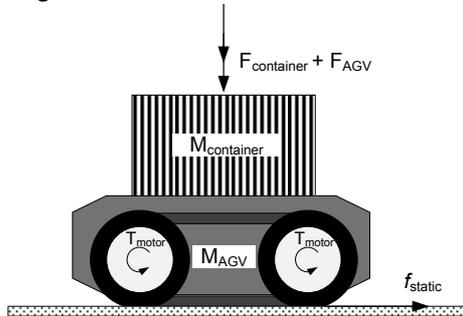

Fig. 3. A model of AGV which is transporting a container

The force which is required to move the AGV can be approximated by the following equations.

$$F_{AGV} = F_{moving} + F_{friction} \qquad (1)$$

where:
$F_{AGV}$ = total force required by the AGV (N)
$F_{moving}$ = force to drive the AGV (N)
$F_{friction}$ = friction that must be resisted (N)

AGV driving force and friction force can be calculated using the following equations.

$$F_{moving} = m \cdot a \qquad (2)$$

$$F_{friction} = c \cdot N \qquad (3)$$

where:
m = total mass of container and AGV (kg)
a = maximum acceleration ($m/s^2$)
c = rolling friction between tire and the road
N = total normal force (N)

$$N = (m_{AGV} + m_{containers}) \cdot g \qquad (4)$$

where :
$m_{AGV}$ = mass of the AGV (kg)
$m_{AGV}$ = mass of the container (kg)
g = gravitation (9.8 $m/s^2$)

Thus, the required motor torque and motor power can be calculated using the following formulae.

$$T = F_{AGV} \: x \: r_{eff.wheel} \qquad (5)$$
$$P = F_{AGV} \: x \: v \qquad (6)$$

where:
T = motor torque requirement (Nm)
$r_{eff.wheel}$ = effective radius of the AGV wheel (m)
P = motor power requirement (watt)
v = maximum speed of AGV (m/s)

Friction coefficient is listed in table 3 [32] and ocean standard containers specification is listed in table 4 [33].

TABLE 3
FRICTION COEFFICIENT [32]

| | Rolling Frictional Coefficient - *c* |
|---|---|
| Truck tire on asphalt | 0.006 - 0.01 |
| Ordinary car tires on concrete | 0.01 - 0.015 |

TABLE 4
OCEAN CONTAINER STANDARD [33]

| Container Size | Container Weight | Maximum Loading | Total |
|---|---|---|---|
| 20' standard | 2,220 kg | 22,100 kg | 24,320 kg |
| 40' standard | 3,740 kg | 27,397 kg | 31,137 kg |
| 40' high cube | 3,950 kg | 29,600 kg | 33,550 kg |
| 45' high cube | 4,470 kg | 28,390 kg | 32,860 kg |

To complete the design data, some assumptions are shown in table 5 based on reference [31].

TABLE 5
DATA ASSUMPTION FOR AGV DESIGN

| AGV Parameter | Assumption Data |
|---|---|
| Maximum speed | 6 m/sec |
| Maximum Acceleration | 2 m/sec² |
| AGV Dead weight | 30,000 kg |
| Effective wheel radius | 0.25 m |

Therefore, a hypothetical specification of AGV design is proposed as shown in table 6.

TABLE 6
THE PROPOSED AGV HYPOTHETIC SPECIFICATION

| Container Type | Electric Motor (Total) Specification | | |
|---|---|---|---|
| | Output Torque (KNm) | Power (KWatt) | Output Speed (rpm) |
| 20' standard | 30 | 700 | |
| 40' standard | 33 | 789 | 228 |
| 40' high cube | 35 | 820 | |
| 45' high cube | 34 | 809 | |

The hypothetic specification of AGV design in table 6 is ideally realized using 4 electric motors attached at each wheel which has specification as follows: 225 KW, 229.3 rpm and 9.37 KNm. However it may be realized using an electric motor which has the following specification: 855 KW, 462 rpm, and 17.7 KNm [34]. To meet the specification compliance, a reducer gearbox with gear ratio of 2 is added.

*Battery Capacity*

One key aspect of vehicle autonomy is power consumption which has become particularly relevant in applications with critically limited power sources [35]. The AGV is equipped with batteries as energy source. As the consequences, it has limited operation time depending on battery capacity. Thus power efficiency is very important to have a long time operation AGV [36]. Table 7 shows several types of battery cells that can be used along with their specifications. To obtain a specification that matches the design of AGV, the battery cell can be arranged in series to obtain the desired working voltage, and in parallel to obtain the desired energy capacity.

TABLE 7
BATTERY CELL SPECIFICATIONS [37]

| Battery Type | Specification per cell | | |
|---|---|---|---|
| | Voltage | Approx. Energy | Safety |
| $LiFePO_4$ | 3.2 V | 120 wh | Safe |
| Lead acid | 2.0 V | 35 wh | Safe |
| NiCd | 1.2 V | 40 wh | Safe |
| NiMH | 1.2 V | 80 wh | Safe |
| $LiMn_xNi_yCo_zO_2$ | 3.7 V | 160 wh | Better than LiCo |
| $LiCoO_2$ | 3.7 V | 200 wh | Unsafe |

B. *The Proposed AGV Control System*

An effective control system is required to make the MAGVS can carry out any task which is assigned by supervisory computer [38]. The proposed control system for AGV in this paper is shown in figure 4. Note that the AGV takes form of 4 wheels independent drive.

The AGVs control system in this paper uses distributed control system method. The control system is divided into several independent parts which includes driver module, motor controller module, motor feedback module, navigation system module, wireless communication module, and vehicle control module. Each module is explained in the following paragraphs.

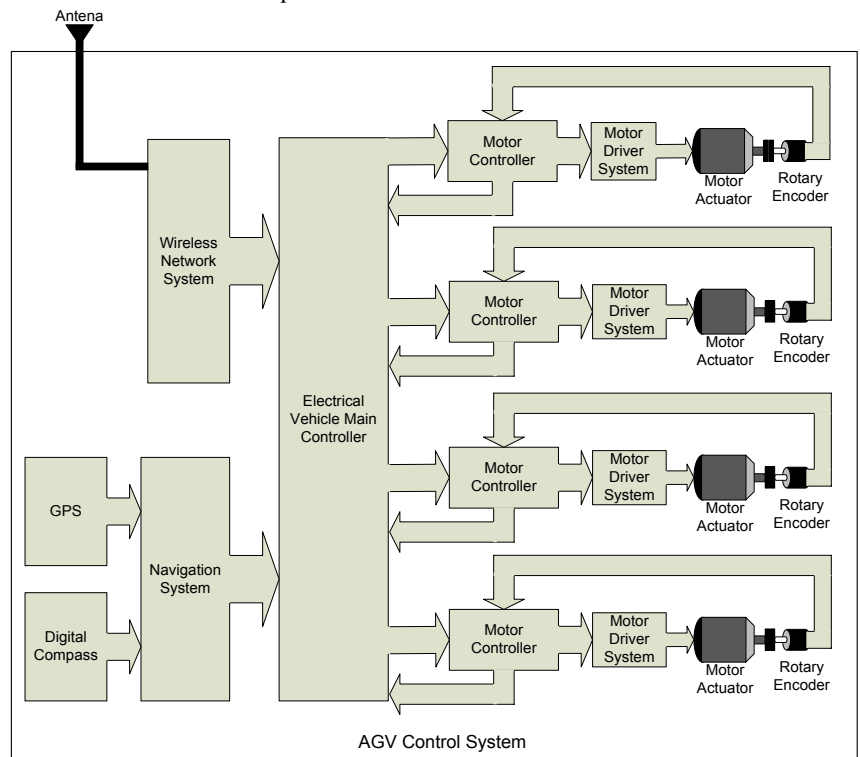

Fig. 4. The proposed control system for AGV system

## Motor Driver

The motor driver receives a signal from the motor controller to drive the motor in the CW or CCW direction, and adjusts the amount of power flowing into the motor.

## Motor Controller

Motor controller system serves to control the movement of each motor actuator. The motor controller module receives commands and reference signals from the vehicle controller containing position and relative speed information to be met by the motor actuator. Motor control module provides a signal to the motor driver module in the form of on / off, motor direction, and motor brake. In addition, the power supplied to the motor driver is also regulated through a PWM generator. Feedback signal received from the rotary encoder is compared with a reference counter to control the vehicle. The motor controller module also updates the relative position data to the vehicle main controller based on the cumulative change in the position of the motor actuator. Figure 5 shows an illustration of block diagram of the motor control module in the AGV system.

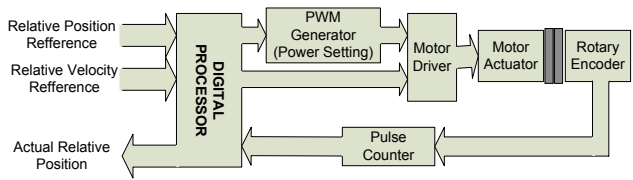

Fig.5. Block diagram of motor control module in the AGV system.

Motor control functions are realized using control algorithm software programmed in a digital processor. Figure 6 shows an example of a simple control algorithm.

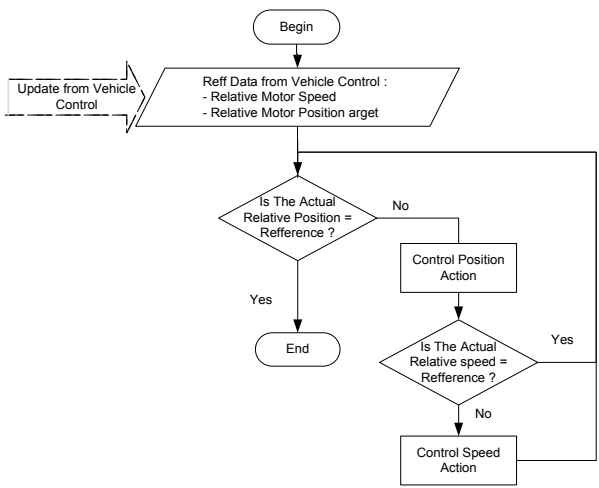

Fig. 6. An algorithm for motor control module

## Wireless Communication System

Wireless module is used to handle communication between the supervisory controller and all AGVs. This wireless communication module sends a command signal from the supervisory controller to each AGV in the form of command mode (e.g.: loading, unloading, standby, moving targets, etc.), reference path to be followed by the AGV, and continuous speed along the path. A communication system is constructed by transmitter and receiver wireless modules. Figure 7 shows a concept of remote control using wireless system between the supervisory computer and the MAGVs.

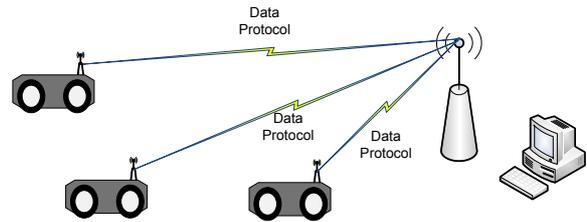

Fig. 7. A concept of remote control using wireless system

## Navigation System

Navigation system is used to guide the movement of the MAGVs from initial position to target position to perform its mission [39]. There are two techniques namely fixed trajectory navigation and free ranging navigation with no fixed path. In the fixed path method, a trajectory guidance, such as magnetic tape, photo reflective tape on the floor or burying of wire below the floor, will guide the AGV in the fixed trajectory. This method, however, leads to a rigid system. In case of the free ranging technique, the AGV has a map of the navigation area and several fixed reference points which can be detected by onboard sensors of the AGV. The AGV localizes itself on the basis of perceived locations of these reference points.

In this paper the free ranging technique is proposed. The free ranging technique allows the AGV move freely without any fixed guidance path so that the AGV operations can be performed more flexible. Figure 8 shows a block diagram illustration of the proposed MAGVs navigation system.

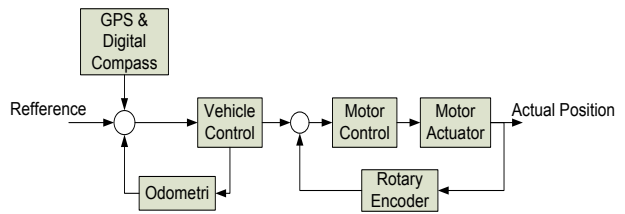

Fig. 8. The proposed MAGVS navigation system.

To measure AGV relative position, rotary encoder sensor is used. AGV mileage is determined from the number of revolutions of each AGV wheel which is estimated from the number of rotary encoder pulses. Motion direction of AGV is determined by rotating direction of each AGV wheel and AGV movement angle.

Moving distance $s$ and turning angle $\vartheta$ of the AGV can be calculated as follows.

$$s = \frac{count\ of\ pulses}{number\ of\ pulses\ per\ revolution} . 2\pi . r_{effective} \quad (7)$$

$$(inside\ arc\ length - outside\ arc\ length) = d . \vartheta \quad (8)$$

where $d$ is the wheel base distance.

The proposed navigation system is also equipped with digital compass and GPS (Global Positioning System). The GPS data shows absolute position of the AGV in his work space, while the digital compass shows the movement direction of the robot against the direction of the polar magnetic compass.

GPS determines global position using satellite constellation which is a system consists of 24 satellites [19]. System orbit satellites are arranged so that every place on earth can receive signals emitted from five to eight satellites [40].

A GPS receiver is attached to the device in the earth, which in this paper is AGV which will receive the encoded signal from the satellite in the form of time and position of the orbit satellite. Time differences received by the receiver indicate the distance between the satellite and the receiver. By entering information concerning distance between the receiver and each satellite as well as information concerning the position of each orbit satellite, latitude and longitude of the receiver can be calculated. The accuracy of commercial GPS systems without any augmentation is approximately 15 meters [41]. To obtain higher accuracy, Differential GPS (DGPS) can be used. In a DGPS a fixed receiver with known position is added as shown in figure 9. Fixed receiver position data received from satellites will be calculated as error value, and it is included in the calculation of the position of the other receivers.

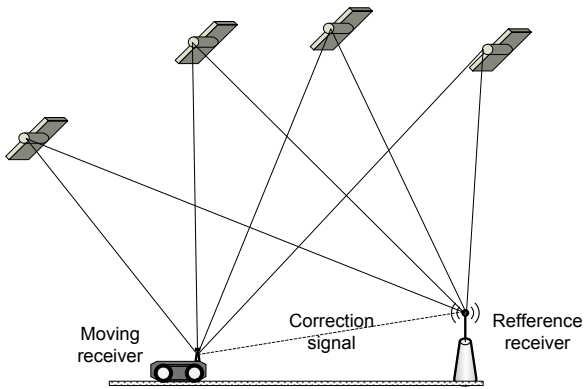

Fig. 9. Illustration of DGPS in the proposed MAGS.

*Electric Vehicle Main Controller*

The vehicle main controller functions to set all modules in the AGV. It communicates with the supervisory computer via the wireless module. Vehicle receives the mode of operation commands protocol control from the supervisor computer such as loading, unloading, standby as well as data path planning, including target position and velocity along the planned path. The received data from the supervisory computer is translated by the vehicle main controller who then gives the command to each motor control module. Electric vehicle main controller also updates the data of AGV position to the supervisor computer as a reference data to coordinate the whole MAGVs.

*C. The Proposed Multi AGV Supervisory Control System*

A supervisory computer is in charge of controlling and coordinating the entire operation of all AGVs in the scope of work. To do that, the main tasks of the supervisory computer cover path planning, coordination and traffic management control.

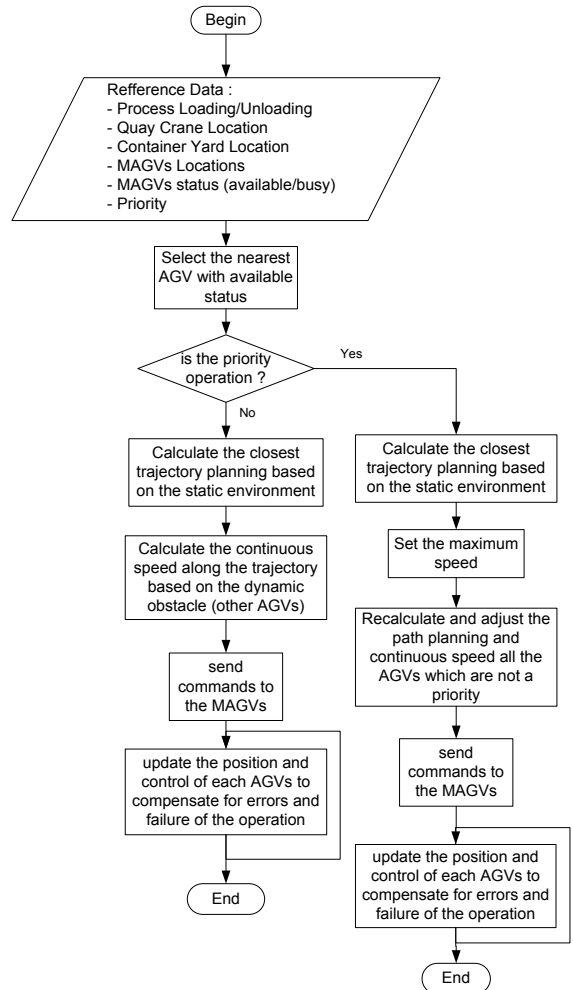

Fig. 10. The proposed simple PBC MAGVS control algorithm.

Path planning control is aimed to determine the route to be taken by all AGVs [42]. The supervisory computer calculates the most optimal path to be followed by MAGVs based on static environments and dynamic environments. After obtaining the proper path planning, supervisory computer sends commands to the AGVs, and then updates the position of each movement of the AGVs so that no collision occurs. In this paper, an algorithm which is a combination of priority method and cooperative method is proposed for determining AGV path planning. Figure 10 shows a simple Priority-Based Cooperative (PBC) MAGVS Control algorithm proposed in this paper.

In this algorithm the default determination of path planning for MAGVS is set by the cooperative method of which the path planning is determined by calculating the nearest trajectory that can be achieved to reach the target point which is obtained by considering the environment and the static obstacles. Speed of each AGV is set continuous along the planned path to avoid a dynamic obstacle in the form of other AGV movement. Speed adjustment is done in sequence where the AGV that will get its mission adjusts its speed along its planned path to avoid collision with AGV that is doing the previous mission. However, if the supervisory computer receives a signal priority for an operation, the priority is given to the AGV which will be conducting that mission with only considers static obstacle and takes the shortest path with maximum speed. Meanwhile, to avoid collision with other AGV, other AGVs which is not given priority re-adjusts its speed to avoid collision.

### D. The Proposed Container Terminal Port Model

Figures 11 to 13 indicate the conditions of containers port terminal in the port harbor of Jakarta Indonesia, Hong Kong port, the Port of Rotterdam, respectively [43]. Harbor port of Jakarta Indonesia is still a conventional containers port having low quality in loading and un-loading services [3]. On the other hand Hong Kong's containers port is one of the best and busiest ports in the world [44]. Container terminal management process has been improved so that even the activity of loading and unloading at the port of Hong Kong is far busier than in the port harbor of Jakarta the waiting time of containers loading and unloading is much faster. Process of transporting of containers in the Hong Kong port still uses conventional container trucks as shown in figure 14 [45].

Port of Rotterdam is one of the major ports in the world which has already been implementing automation technology in the process of containers transportation. This port uses AGVs to transport containers which are guided by the navigation devices in the form of electromagnetic transponders which are embedded in the terminal pavement as shown in figure 15 [46].

In conventional AGV based containers transportation system AGV operates with a guide device which is embedded in the pavement port so that the movement of AGV is rigid and limited. Figure 16 shows conventional AGV based containers transportation system in a port terminal.

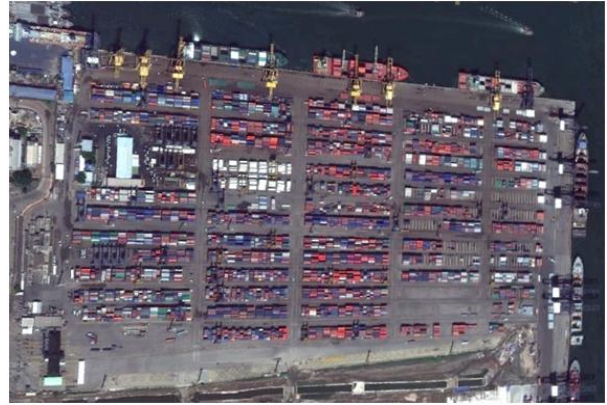

Fig. 11. Tanjung Priok Port Containers Terminal, Indonesia [43]

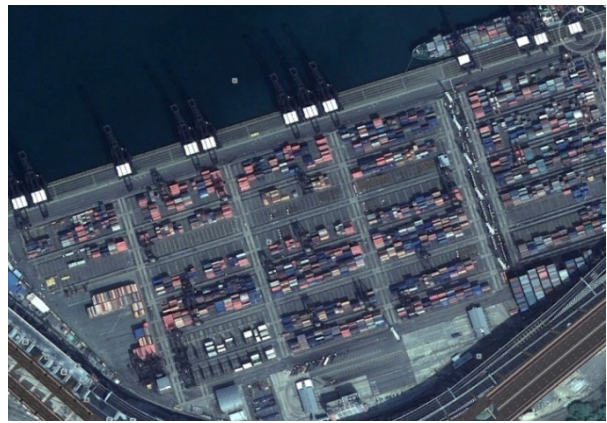

Fig. 12. Hongkong Port Containers Terminal, Hongkong [43]

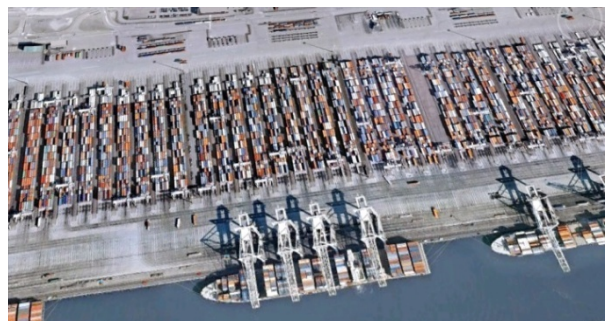

Fig. 13. Rotterdam Port Containers Terminal, Netherland [43]

This paper proposes MAGVS to be applied in port container terminals having free ranging navigation system where trajectory planning is more flexible to increase efficiency and flexibility. Figure 17 exhibits the proposed MAGVS model implemented in a containers terminal.

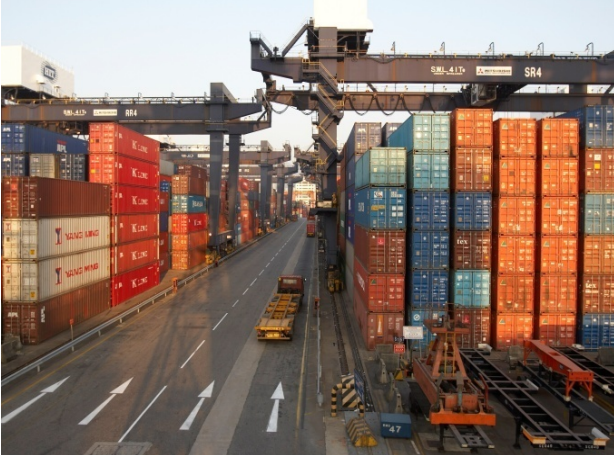

Fig. 14. Hong Kong Port Containers Transportation System [45]

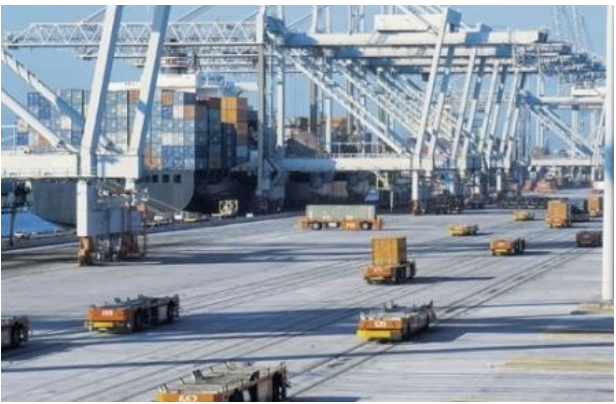

Fig. 15. Rotterdam Port Containers Transportation System [46]

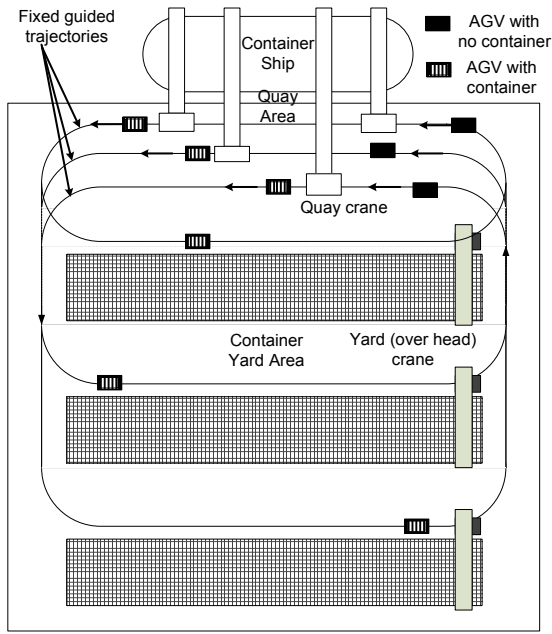

Fig. 16. Conventional AGV based containers transportation system.

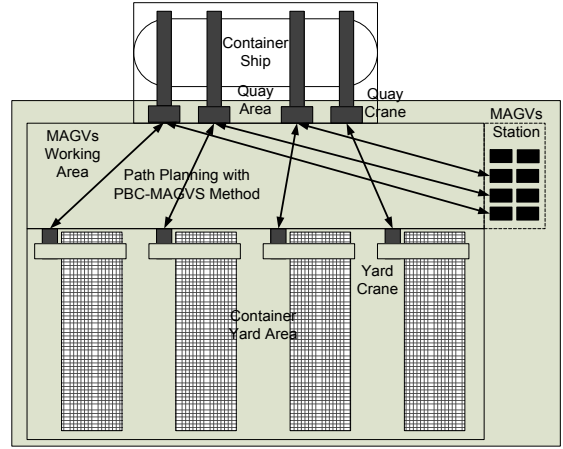

Fig. 17. The proposed MAGVS model in a containers terminal

## IV. CONCLUSION

AGV system with fixed trajectories has been implemented at Rotterdam port where the AGVs are driven by diesel engines. From power analysis result, to be able to transport ocean containers, an electric AGV design specification has been obtained as follows: electric motors total power between 700-900KW and speed of 228 rpm. In order to conduct logistics work at containers terminal a multi AGVs system (MAGVS) can be developed. This MAGVS is better equipped with free ranging navigation system using GPS, digital compass and rotary encoder sensors. Path planning, coordination and traffic management control is conducted by the supervisory controller based on priority-based coordinative (PBC) control algorithm. A model of MAGVS has been proposed using the above mentioned navigation system and PBC control algorithm in order to enhance containers loading-unloading process efficiency.